\documentclass[letterpaper, 10 pt, conference]{ieeeconf}  % Comment this line out if you need a4paper

\IEEEoverridecommandlockouts                              % This command is only needed if 
\usepackage{booktabs}
\usepackage{array,multirow,graphicx}
\usepackage{float}
\usepackage{tabularx}
\usepackage{adjustbox}
\usepackage{amssymb}
\usepackage{pifont}% http://ctan.org/pkg/pifont
\usepackage{wrapfig}
\usepackage{caption}
\usepackage{amsmath}
\newcommand{\cmark}{\ding{51}}
\newcommand{\xmark}{\ding{55}}
\usepackage[table, dvipsnames]{xcolor}

\usepackage[utf8]{inputenc} % allow utf-8 input
\usepackage[T1]{fontenc}    % use 8-bit T1 fonts
\usepackage{hyperref}       % hyperlinks
\usepackage{url}            % simple URL typesetting
\usepackage{booktabs}       % professional-quality tables
\usepackage{amsfonts}       % blackboard math symbols
\usepackage{nicefrac}       % compact symbols for 1/2, etc.
\usepackage{microtype}      % microtypography
\usepackage{xcolor}         % colors
\usepackage{graphicx}
\usepackage{multirow}
\usepackage{wrapfig}

\usepackage{listings}
\usepackage{bibentry}
\usepackage{arydshln}
\usepackage{pifont}

\usepackage[capitalize]{cleveref}

\title{\LARGE \bf
Refining Pre-Trained Motion Models
}

\author{Xinglong Sun$^{1}$, Adam W. Harley$^{1}$, and Leonidas J. Guibas$^{1}$% <-this % stops a space
\thanks{$^{1}$Stanford University, Computer Science Department. {\tt\small \{xs15, harleya, guibas\}@stanford.edu}}
}

\begin{document}

\maketitle
\thispagestyle{empty}
\pagestyle{empty}

\begin{abstract}

Given the difficulty of manually annotating motion in video, the current best motion estimation methods are trained with synthetic data, and therefore struggle somewhat due to a train/test gap. Self-supervised methods hold the promise of training directly on real video, but typically perform worse. These include methods trained with warp error (i.e., color constancy) combined with smoothness terms, and methods that encourage cycle-consistency in the estimates (i.e., tracking backwards should yield the opposite trajectory as tracking forwards). In this work, we take on the challenge of improving state-of-the-art supervised models with self-supervised training. We find that when the initialization is supervised weights, most existing self-supervision techniques actually make performance worse instead of better, which suggests that the benefit of seeing the new data is overshadowed by the noise in the training signal. Focusing on obtaining a ``clean'' training signal from real-world unlabelled video, we propose to separate label-making and training into two distinct stages. In the first stage, we use the pre-trained model to estimate motion in a video, and then select the subset of motion estimates which we can verify with cycle-consistency. This produces a sparse but accurate pseudo-labelling of the video. In the second stage, we fine-tune the model to reproduce these outputs, while also applying augmentations on the input. We complement this boot-strapping method with simple techniques that densify and re-balance the pseudo-labels, ensuring that we do not merely train on ``easy'' tracks. We show that our method yields reliable gains over fully-supervised methods in real videos, for both short-term (flow-based) and long-range (multi-frame) pixel tracking. Our code can be found here: \url{https://github.com/AlexSunNik/refining-motion-code}.

\end{abstract}

\section{Introduction}

Given the difficulty of manually annotating motion in video, the current best motion estimation methods are trained with synthetic data~\cite{flownet,flyingthings16,sintel,wang2020tartanair,zheng2023point}. 
%Highly diverse synthetic  is the case in optical flow~\cite{teed2020raft}, multi-frame point tracking~\cite{harley2022particle}, and even recent SLAM methods~\cite{teed2021droid}. 
This synthetic training data is diverse, but highly unrealistic, forcing the methods to contend with a sim-to-real gap. 
%and must contend with a sim-to-real gap. %This is especially true for fine-grained motion problems, such as optical flow~\cite{teed2020raft} and multi-frame point tracking~\cite{harley2022particle}.

Self-supervised (or ``unsupervised'') models hold the promise of training directly on real video, yet lag behind their supervised counterparts. State-of-the-art self-supervised motion methods typically rely on warp error (i.e., color constancy) combined with smoothness terms~\cite{yu2016back,selflow,stone2021smurf}, and cycle-consistency (i.e., tracking backwards should yield the opposite trajectory as tracking forwards)~\cite{wang2019unsupervised,li2019joint,wang2019learning,jabri2020walk}. In current experimental setups, fully-supervised versions are typically positioned in a league ahead, reflecting in some way an unfair advantage. This divide between supervised and unsupervised methods is purely academic; if synthetic supervision is available and helpful, it makes sense to use it.

 In this work, we improve state-of-the-art supervised models with self-supervised training in an unabelled test domain. 
 Unlike typical self- or semi-supervised motion estimation approaches which are trained from scratch, we refine an off-the-shelf pre-trained model for a given test domain without requiring any additional data.

 %for any test domain. 
 %This regime may be broadly called ``semi-supervised'', but our specific setup is to take an arbitrary pre-trained model off the shelf, and attempt to improve it for a test domain---even without accessing the data the model was trained on. 
  %With the hope to simultaneously tackle both issues (\textbf{A and B}), 
 %In this work, we take on the challenge of improving state-of-the-art supervised models with self-supervised training in an arbitrary unlabelled test domain. Unlike typical self-supervised motion estimation approaches, which start from scratch, our setup is to take an 
 %arbitrary pre-trained model off the shelf, and attempt to improve it for any test domain. 
 %We argue that this setup reflects a very practical and common use-case: a user observes undesired motion estimation results when applying the provided pre-trained model to user-given videos, and is willing to invest some GPU cycles into enhancing the model performance. 
 We argue that this setup reflects a very practical use-case: a user notices some errors in the model's output, and is willing to invest some GPU cycles into tuning up the results---provided that the tuning is effective.

In this setup, the most straightforward solution is to directly optimize standard self-supervision objectives in the unlabelled data. We find that this reliably makes the pre-trained model perform even worse, 
%to directly apply standard self-supervision losses on the pre-trained model with videos from the target domain. However, we observe that it reliably makes the pre-trained model even worse, 
suggesting that the self-supervised training signals are too imprecise and noisy to provide useful gradients to an already well-optimized motion model. This observation is consistent with the fact that self-supervised models perform worse than supervised ones in general. 
%Consistent with the fact that self-supervised models perform worse in general, we find that applying standard self-supervision losses on a pre-trained model reliably makes it worse. It appears that the training signal is too imprecise to provide useful gradients to models in this context.

In order to obtain a ``clean'' training signal from real-world unlabelled video, we propose to separate labeling and training into two distinct stages. In the first stage, we use the pre-trained model to estimate motion in a video, and then select the subset of motion estimates which we can verify with cycle-consistency. This produces a sparse but accurate pseudo-labeling of the video. In the second stage, we fine-tune the model to reproduce these outputs, while also applying augmentations on the input. We complement this self-training method with simple techniques that densify and re-balance the pseudo-labels, ensuring that we do not merely train on ``easy'' tracks. 

In our experiments, we demonstrate that our method yields reliable gains over fully-supervised methods in real videos. Our experiments cover two types of motion models: optical flow (i.e., RAFT~\cite{teed2020raft}), and long-term point tracking (i.e., PIPs~\cite{harley2022particle}). We use standard benchmarks for the analysis: MPI-Sintel~\cite{sintel} for flow, and CroHD~\cite{sundararaman2021tracking}, Horse30~\cite{mathis2021pretraining}, and TAP-Vid-DAVIS~\cite{doersch2022tap} for point tracking. Despite starting from pre-trained state-of-the-art weights, our self-supervision framework produces consistent improvements in results. %, whether the optimization is performed per-video or 
%consistent improvements over the initial state-of-the-art weights. 
We hope that our technique, and our semi-supervised setup, will inspire future work into refining pre-trained motion models.

\section{Related  Work} \label{sec:related}

\textbf{Learning motion from color and smoothness.}
Soon after the introduction of deep-learned flow methods~\cite{flownet}, there has been interest in learning optical flows in self-supervised manner~\cite{yu2016back}, converting classic flow assumptions~\cite{lucas1981iterative}, color constancy, and motion smoothness, into supervision objectives. 
Color constancy implies that when flow is accurate, corresponding pixels in consecutive frames should have the same color. By warping images to align with estimated flow, reducing per-pixel differences can improve flow accuracy, particularly when computed across multiple scales ~\cite{burt1987laplacian}. The smoothness assumption helps to resolve ambiguities and occlusions which can be simply implemented by penalizing spatial gradients on the output flow. Minute details in this setup can affect performance greatly~\cite{jonschkowski2020matters}. Instead of relying on color and smoothness, which can be understood as indirectly encouraging correct motion, we compute pseudo-labels for unlabelled data, providing direct regression targets.  %which is intended to provide always-correct regression targets. 

\textbf{Bootstrapping with student-teacher setups.} 
%Together with color matching, 
%with optimizing the details of this setup~\cite{jonschkowski2020matters},
Some recent works stabilize the losses from color-matching by using student-teacher setups. In this setting, the teacher is a moving average copy of the student, and the student receives harder data than the teacher~\cite{selflow,stone2021smurf,im2022semi}. Our work can also be considered a student-teacher setup, but the teacher is a frozen copy of a pre-trained model, and the student is a trainable version of the same model. Our setup is in line with knowledge distillation~\cite{hinton2015distilling}, except our pair of models begins with the same architecture and weights (rather than the student being a smaller model). 
% model to another, we transfer behavior from the model to itself, focusing on specific samples where the model was deemed successful. % to transfer select behaviors from the model to itself,  reinforce certain behaviors and penalize others
%certain behaviors within the model. %behavior that already exists in the model. % the model \textit{to itself}, 

\textbf{Learning motion from cycle-consistency.}
Color matching is known to break down under some conditions (e.g., specularities and occlusions), so many works focus instead on learning directly from cycle-consistency~\cite{wang2019unsupervised,li2019joint,wang2019learning}. The key idea is that after tracking a target from a given startpoint to an estimated endpopint, reversing the video and re-applying the tracker from the endpoint should lead back to the original startpoint. This core idea is typically combined with patch-level affinity matrices, which can be traversed with spatial transformers~\cite{wang2019learning}, region-level motion averages~\cite{li2019joint}, or random walks~\cite{jabri2020walk,tang2021breaking,bian2022learning}. Unlike existing self-supervision works which initialize weights from ImageNet~\cite{imagenet_cvpr09} or randomly, our method begins with state-of-the-art models whose architectures and weights are pre-optimized for motion estimation~\cite{teed2020raft,harley2022particle} and attempts to further refine them on a new test domain. Compared to previous works which directly leverage cycle consistency as part of training objective~\cite{wang2019learning}, we use it to design a filter to select cycle-consistent motion estimates for the finetuning.
% In the context of our work, the architecture and initialization are key details: existing works invent new architectures intended to make self-supervision easier, and weights are either initialized from ImageNet~\cite{imagenet_cvpr09} or from random. Our work instead begins with state-of-the-art models whose architectures and weights are pre-optimized for motion estimation~\cite{teed2020raft,harley2022particle}, which forces the self-supervision method to be architecture-invariant and focuses the problem on fine-tuning rather than learning from scratch. Our method also differs in the way in which cycle-consistency is used: rather than penalize estimates  which are not cycle-consistent, we first gather cycle-consistent estimates and then encourage the model to reproduce these estimates for augmented versions of the input. 
% which essentially forces the self-supervision design to be architecture-invariant. 
%\cite{xu2021rethinking}
%\cite{tang2021breaking}

%with a bilinear resampling, or warp, which uses the estimated flow to bring the second image into alignment with the first image; the objective is to reduce the absolute difference between this warped image and the original first image. 

\textbf{Semi-supervised motion estimation.} 
Semi-supervised methods in motion estimation typically exploit a mix of labeled and unlabelled data, hoping to achieve better performance than exclusively using labeled data. 
Lai et al.~\cite{lai2017semi} propose to learn a discriminator on warp errors from ground truth warps to provide feedback on the overall quality of the warps,replacing the noisy per-pixel color cues. 
Jeong et al.~\cite{jeong2022imposing} add a separate segmentation module to a flow network, supervise this module with occlusion masks applied onto unlabelled images, and also ask flows to be consistent across transformations applied to the input, which is similar to our goal but requires modified architecture for segmentation. %Closer to our work, Zhang et al.~\cite{zhang2022clip} use the RAFT optical flow model~\cite{teed2020raft} and attempt to improve it, though again relying on supervised synthetic data at the same time as unlabelled data. 
% Zhang et al.~\cite{zhang2022clip} propose to alternate between supervised and self-supervised learning with a contrastive learning objective and student-teacher bootstrapping process, which is similar to ours but operates with dense flow maps compared to our sparse but accurate pseudo-labels filtered with cycle-consistency.

\textbf{Training from pseudo labels.}
The technique of training on self-generated estimates (i.e., pseudo labels) was first proposed by Lee et al.~\cite{lee2013pseudo}. %, in which confident estimates are treated as training targets for fine-tuning. 
This core idea is often paired with methods to filter the training targets to ``confident'' ones, where confidence may be approximated with the help of ensemble methods~\cite{10.5555/2898607.2898816,chen2013neil}. Our work computes confidence with the help of domain knowledge: if a trajectory is within some cycle-consistency margin and color constancy, we consider it a pseudo-label. Similar to Chen et al.~\cite{chen2013neil}, we assume that it is possible to train on a small number of confident samples and generalize to other regions of the domain. 

\section{Method}
%As mentioned, 

Our method begins with a pre-trained motion estimator and unlabelled videos from test domain, and attempts to improve the estimator performance with self-supervision on the input videos. % with self-supervision in unlabelled video.
%by first gathering pseudo-labels, and then refining the model to reproduce those estimates in more challenging settings. 
Our overall framework, illustrated in Fig.~\ref{fig:method}, consists of two stages. 
In the first stage, we generate pseudo-labels, by running the pre-trained model on the videos and selecting a subset of model estimates with cycle-consistency-based filters. 
%In the first stage, we use the pre-trained model to generate motion estimates.
%We then select the accurate subset of these estimates, using cycle-consistency-based filters, and treat these as pseudo-labels. 
In the second stage, we refine the model by fine-tuning it on those pseudo-labels, challenging it to reproduce those estimates under augmentations. % In this section, we first describe the motion models in consideration then step through the method.
In this section, we step through the method: we first describe our motion models, then discuss pseudo-labeling, and end with refinement. 
%Accurate estimates are then selected by our designed cycle-consistency-based filters and used as pseudo-labels. In the second stage. we finetune the pre-trained model to reproduce those estimates in more challenging settings. In this section, we first describe the motion models in consideration then step through the method.

% Our method begins with pre-trained motion estimators, and attempts to improve them with self-supervision in unlabelled video, by first gathering pseudo-labels, and then refining the model to reproduce those estimates in more challenging settings. In this section, we step through the method. We begin by describing our motion models, then discuss pseudo-labeling, and end on refinement. 
\subsection{Preliminaries: RAFT and PIPs}

Our approach can be applied to any motion model, but we focus here on two popular methods: % motion models, but we focus on two 
%As our base motion models, we use 
RAFT~\cite{teed2020raft} and PIPs~\cite{harley2022particle}. These are fine-grained general-purpose motion models, in the sense that they can track arbitrary small elements (i.e., points on the scene surface, specified by an $(x,y)$ coordinate in a given frame) in arbitrary videos (i.e., not restricted to particular domains, though potentially suffering a generalization gap). We use pre-trained weights publicly released by the authors, which were produced by training in synthetic data. In general, any motion model could be used, but we choose RAFT and PIPs as they represent the state-of-the-art for optical flow and multi-frame point tracking.  

% The flow map is a spatial displacement map, indicating the frame-1 correspondence for each pixel in frame-0. RAFT works by iteratively refining an estimated flow map, using feature correlations indexed in a local window around the current estimates.
RAFT is a \emph{\textbf{2-frame dense-motion model}}. It consumes two consecutive RGB frames as input, and produces a dense flow map as output. 
The flow map is a spatial displacement map, indicating the frame-2 correspondence for each pixel in frame-1. 
%Optical flow can be linked across time to form longer-range correspondences, but are hindered by the sensitivity to occlusion and slow computation. %There exist RAFT checkpoints trained on various combinations of datasets~\cite{raftcode}; we use a checkpoint trained on FlyingThings
Optical flow can be linked across time to form longer-range correspondences, but these trajectories are sensitive to occlusions, and slow to compute. 
RAFT works by iteratively refining an estimated flow map, using feature correlations indexed in a local window around the current estimates. 

% PIPs' design is inspired by RAFT, and similar works by iteratively refining its estimates, using feature correlations indexed in a local window around the current estimates. 
PIPs is an \emph{\textbf{8-frame sparse-motion model}}. It consumes a sequence of 8 consecutive frames as input, along with a list of target coordinates to track, and it produces an 8-frame trajectory for each target. These 8-frame trajectories are less sensitive to occlusions than optical flow, but are typically sparse since computing these trajectories densely is expensive. 
PIPs' design is inspired by RAFT, and similarly operates by iteratively refining its estimates, using feature correlations indexed in a local window around the current estimates.

% The key component here is our cycle-consistency-based filter which is used to collect the subset of accurate and clean motion estimates produced by the pre-trained model. 
% The key component here is how we design this filter to keep 'good' and clean estimates.
\subsection{First Stage: Pseudo-label Generation}
In the first stage, we generate pseudo-labels, by applying the pre-trained model to unlabeled videos and collecting a reliable subset of the motion estimates, where reliability is estimated with cycle-consistency. % with a cycle-consistency-based filter. Measuring cycle consistency requires running the motion model on video inputs in both forward and backward directions, and also requires bringing the estimates into alignment (e.g., via warping) to enable comparison.

% Cycle-consistency is measured as the distance between the start point (on frame 0) and the final endpoint (also at frame 0). If this distance is below a threshold, we call the track cycle-consistent. 
Cycle-consistency leverages the fact that if we track forwards in time without error, then start at the predicted endpoint and track backward in time without error, we will end up at the original startpoint. Errors in tracking typically break this cycle. 
%For 2-frame optical flow (RAFT) and 8-frame tracking (PIPs), we define the cycle consistency a bit differently since typical errors are at different magnitudes for various timescales and motion densities. 
We define the cycle consistency similarly for 2-frame optical flow (RAFT) and 8-frame tracking (PIPs). %,  typical errors are at different magnitudes for various timescales and motion densities.

\textbf{2-Frame Cycle-Consistency Filter.}
For optical flow, we compute consistency by warping the backward flow map into alignment with the forward flow map and then measure per-pixel disagreements with a sum (since opposite flows should cancel out)~\cite{brox_densepoint}. Denoting the forward flow as $w$ and the aligned backward flow as $\hat{w}$, $\forall$ pixel locations $i,j$, we check: 
% \begin{equation}
% \label{eqn:raftconsty}
%     || w + \hat{w} ||_2^2 < \alpha ( ||w||_2^2 + ||\hat{w}||_2^2 + \beta ), 
% \end{equation}
\begin{equation}
\label{eqn:raftconsty}
    || w_{ij} + \hat{w}_{ij} ||_2^2 < \alpha ( ||w_{ij}||_2^2 + ||\hat{w}_{ij}||_2^2 + \beta ), 
\end{equation}
which linearly scales the tolerance threshold according to the motion magnitude. Flows that satisfy Eqn.~\ref{eqn:raftconsty} are considered cycle-consistent. 

% We use $\alpha=0.005$ and $\beta=0.25$, which is more strict than prior work has used, for example to estimate occlusion~\cite{brox_densepoint,xu2022gmflow}.
% The reason for this is that we would prefer to discard an estimate if we are not highly confident that it is correct; sparse but accurate pseudo-labels are better than denser but noisier labels. 
%The reason for this is that we seek to train the model only on 
% We use the flow magnitude to create an adaptive threshold for this disagreement map. 
%\begin{equation}
%    \textrm{Warp}(F_{2\rightarrow 1},  )
%\end{equation}
%Since errors are in a different magnitude for 2-frame vs. 8-frame tracking, we choose different 
% We note that we use this a \textit{filtering} mechanism for pseudo ground truth, while prior work has used it directly as a supervision target

Following the classical optical flow assumption of color constancy~\cite{lucas1981iterative, yu2016back, jonschkowski2020matters}, we also ask for the tracks to be color-consistent. % augment the filter with color consistency. 
Denoting the zeroth frame as $r$ and the flow-aligned first frame as $\hat{r}$, we use:
\begin{equation}
\label{eqn:raftcolorconsty}
    % || r - \hat{r} ||_2^2 < 0.1, 
    || r - \hat{r} ||_2^2 < \gamma, 
\end{equation}
where the image colors are normalized to the range $[-0.5, 0.5]$. This filter demands that corresponding pixels should have similar colors. Only tracks that are both color and cycle-consistent will be kept by the filter.

% follow Doersch et al.~\cite{doersch2022tap}, and compute cycle-consistency distances in a resolution-normalized space: we first resize the video to $256 \times 256$ (and appropriately rescale the trajectories), and then This is again a very strict threshold compared to accuracy metrics used in existing work~\cite{doersch2022tap}; our selection of these threshold parameters was guided by early experiments in labelled synthetic data~\cite{flyingthings16}.
\textbf{8-Frame Cycle-Consistency Filter.}
For 8-frame sparse motion tracking, we measure the maximum per-timestep distance between the forward trajectory $v$ and the time-flipped backward trajectory $\hat{v}$: 
\begin{equation}
    \label{eqn:consty_metric}
    \max_t || v_t - \hat{v_t} ||_2^2 < \tau, 
\end{equation}
where a threshold of $\tau = 1$ means that the trajectories may only drift a maximum of 1 pixel apart. Tracks that satisfy Eqn.~\ref{eqn:consty_metric} are considered cycle-consistent.

For both 2-frame and 8-frame tracking, the threshold values ($\tau$, $\alpha$, and $\beta$) depend on meta information such as  the spatial resolution of the video. % and we show in our ablation studies that optimal threshold values exist for a given setting. 
We measure the effect of these thresholds in our ablation study, and find that there is a Goldilocks zone: with thresholds set too small, yielding a strict filter, there are too few good tracks for the pre-trained model to finetune; with thresholds set too large, yielding a loose filter, there are too many noisy tracks which corrupt the pre-trained weights.

We run the motion model with temporal stride 2 (to reduce data redundancy), and use our cycle-consistency filters to select a small subset (approximately 10\%) of the resulting tracks. We use this subset as pseudo-labels for the next stage. 
%in the unlabelled video data from target test domain, and export the filtered tracks, producing pseudo-labels for the next stage. % used as pseudo-labels for the next stage. 

% This process creates pseudo-labels for approximately 20\% of the data in the case of RAFT, and approximately 5\% of the data in the case of PIPs. 
%The idea is to track a target forwards, and then, starting at the endpoint of the trajectory, track backwards, and measure how closely we complete a cycle. 

\subsection{Second Stage: Motion Refinement}
\label{subsec:motionref}
In the second stage, we try to improve the pre-trained motion model by fine-tuning it on the batched and augmented pseudo labels, challenging the model to reproduce those motion estimates under more difficult settings. It may seem strange to fine-tune the model using its own estimates, but there are several reasons this may be helpful. 
%\begin{itemize}
    %\item 
    First, the model is iterative (in the case of both RAFT and PIPs), and we are fine-tuning \textit{all} iterations with the model's \textit{final} iteration, making the early iterations more confident to jump toward the answer.     Second, taking gradient steps on these pseudo-labels may lead the model to ``forget'' behavior which is inconsistent with the pseudo-labels, which is a form of domain adaptation. 
    Third, we apply augmentations while fine-tuning, which widens the domain where the model should produce these cycle-consistent outputs. 
%\end{itemize}
We use the default augmentations from RAFT and PIPs, which include color jittering, cropping, resizing, and random occlusions. 

\begin{figure}
\centering
\includegraphics[width=\linewidth]{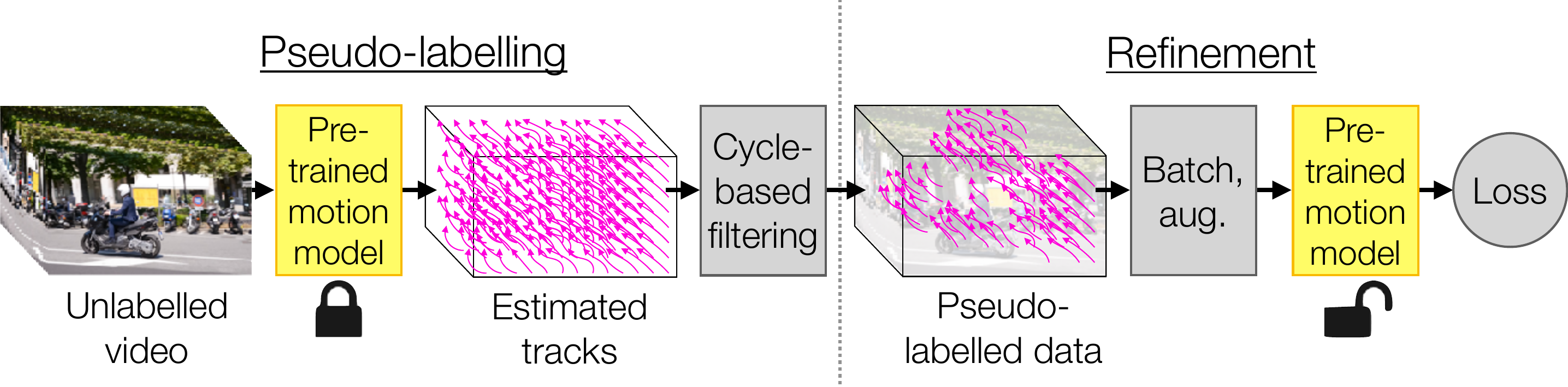}
\caption{\textbf{Refining a pre-trained motion model.} We apply a pre-trained motion model on an unlabelled video, yielding a dense set of tracks. We filter this down to a sparse set of cycle-consistent tracks, creating pseudo-labels. We then train on the pseudo-labelled data using augmentations, to improve the model. % on the video. 
}
\label{fig:method}
%\vspace{-15pt}
\end{figure}

If the test domain is a single video, we simply fine-tune on it directly, as shown in Fig.~\ref{fig:method}. If the test domain has multiple videos, we find that the algorithm is robust to either mixing all video clips together, or fine-tuning in an online fashion, continuing the optimization for each video in sequence. 
%We also propose two versions of fine-tuning settings, designed for single and multiple videos based on the use case. For \textbf{\emph{single-video finetuning}}, we simply perform the described procedure outlined in Fig. \ref{fig:method}. For \textbf{\emph{multiple-video finetuning}}, it is similar, but instead of starting from the pre-trained motion model, we start from the model obtained by finetuning previous videos to benefit from previous refinements. Obviously, multiple-video finetuning is beneficial to a video dataset with many similar samples but would fail if each individual video differs too much.

An important hyperparameter in our approach is the number finetuning iterations $\kappa$, which controls how long the model is trained on the pseudo-labels. In the ablation study we measure the performance at different settings of $\kappa$, and find that as $\kappa$ goes beyond its optimal value, performance degrades slightly but does not collapse. 
%We show in the ablation study that an optimal value of $\kappa$ does exist for a given setting: %with $\kappa$ too small, model does not move from its original weights; 
%with $\kappa$ too large, the model does not collapse, but is slightly worse than the optimal value. %; however, with $\kappa$ too large, model will overfit on the pseudo-labels.

% To not "forget" too much from its training distribution, we only take 100 gradient steps with the pseudo-labels. In order to reduce noise in the optimization, we accumulate gradients across a batch size of 16 for each gradient step. 

%seek to train the model to reproduce these estimates in slightly wider settings. That is, we oul that we are fine-tuning produced 

%\begin{figure}
%  \centering
%  \fbox{\rule[-.5cm]{0cm}{4cm} \rule[-.5cm]{4cm}{0cm}}
%  \caption{Sample figure caption.}
%\end{figure}

\section{Experiments}\label{sec:exp}

%We conducted extensive experiments to validate and demonstrate the effectiveness of our approach 
We present experiments on three widely-used datasets: CroHD~\cite{sundararaman2021tracking}, Horse30~\cite{mathis2021pretraining}, Tap-Vid-DAVIS~\cite{davis2017,doersch2022tap}, and MPI-Sintel~\cite{sintel}. We use Sintel to measure performance on 2-frame dense optical flow estimation, and use CroHD and Tap-Vid-DAVIS to measure performance on 8-frame sparse motion tracking. We use RAFT~\cite{teed2020raft} for optical flow, and PIPs~\cite{harley2022particle} for multi-frame tracking. We use the publicly released checkpoints from the authors. Both models are (pre-)trained on the synthetic FlyingThings~\cite{flyingthings16} dataset, which is noticeably unrealistic but produces strong models. 
%which is an unrealistic but diverse synthetic dataset, yielding state-of-the-art performance on the test sets, but  \emph{\textbf{completely different from the three considered datasets}}. 
%This allows us to evaluate our self-supervised motion refinement framework. 
In addition to evaluating the effect of our fine-tuning scheme, we evaluate alternative optimization objectives and also additional supervised and self-supervised baselines, contextualizing our results with existing works. %supervised and self-supervised works. %We demonstrate later that we could further improve the motion estimation of the state-of-the-art pre-trained motion model on the target test domain.
%for optical flow. 
% which cover a wide range of domains, 
%with two different motion models~\cite{harley2022particle, teed2020raft} covering a wide range of domains and both short-range dense optical flows and long-range point tracking.

% \subsection{Implementation details}

% For PIPs~\cite{harley2022particle}, we use the publicly released checkpoint, which is pre-trained on FlyingThings~\cite{flyingthings16}. RAFT~\cite{teed2020raft} has multiple checkpoints available; we use the one trained on FlyingThings~\cite{flyingthings16}, as opposed to the one additionally trained on MPI-Sintel~\cite{sintel}, because we wish to use MPI-Sintel for evaluation. 

\subsection{Datasets}
\label{subsec:dataset}
%For diversity, we conducted experiments on three datasets namely CroHD~\cite{sundararaman2021tracking}, Tap-Vid-Davis~\cite{doersch2022tap, davis2017}, and Sintel~\cite{sintel} each focusing on different scenarios and motion type. In the following paragraphs, we now cover more detail of each dataset.
In this section we describe each dataset in detail. 

\textbf{CroHD~\cite{sundararaman2021tracking}.} This dataset focuses on pedestrian head-tracking in long videos of crowded environments. Ground truth labels are provided on the train-split of the videos, thus we perform our self-supervised refinement framework on pre-trained PIPs~\cite{harley2022particle} on these videos then validate using the ground truth labels. Videos from CroHD are very similar to each other, because all demonstrate pedestrians walking either indoors or outdoors. In this dataset, we fine-tune on multiple videos, continuing optimization from one video to the next. %This suggests a use case of our \textbf{multiple-video finetuning} setting mentioned in ~\ref{subsec:motionref}.

\textbf{Horse30~\cite{mathis2021pretraining}} is a dataset focused on animal pose estimation, specifically targeting Thoroughbred horses. It encompasses 30 varied horses, with 22 body parts labeled over 8,114 frames by an expert. Since its pose is collected in video sequences, we could also leverage the labels to evaluate tracking performance. We focus on multi-frame tracking scenario and employs PIPs~\cite{harley2022particle} as the baseline motion model to refine. Moreover, we fine-tune on multiple videos in this dataset, continuing optimization from one video to the next.

\textbf{Tap-Vid-DAVIS~\cite{doersch2022tap}.} This is a point tracking dataset  %recently collected and benchmarked 
based on the DAVIS 2017 validation set~\cite{davis2017}. The dataset contains 30 challenging videos featuring different scenes and objects ranging from people bicycling to horse jumping captured with different camera motion. Each video is quite short: around $5$ seconds on average. Ground truth labels are provided for approximately $26$ points per video. We refine PIPs~\cite{harley2022particle} on these videos then validate using the sparse ground truth. In this dataset we fine-tune on each video individually, since each video is unique. %Different individual videos here suggest a use case of our \textbf{single-video finetuning} setting.
% This is a very challenging task for our proposed method because our pseudo-track sampling stage could simply miss regions near the query points. Also, due to the short length of the video, there are not many tracks we could sample and exploit. Unlike CroHD, due to the large inter-dataset variance, we consider each video separately in pseudo-track sampling and refinement stages and compare it with the pre-trained model on that particular video or scene for validation.

\textbf{MPI-Sintel~\cite{sintel}.} This dataset consists of 23 animated scenes, with per-pixel ground truth flow. The videos vary widely in speed of motion, occlusion ratio, and scene contents. Two versions of the data are provided (`clean' and `final'), which vary in postprocessing effects (e.g., motion blur). In this data we refine pre-trained RAFT~\cite{teed2020raft}, and validate using the ground truth. 
In this dataset we fine-tune on each video individually, since each video is unique. %Similar to Tap-Vid-DAVIS, the difference in the individual scenes here suggests a use case of our \textbf{single-video finetuning} setting.

\subsection{Evaluation Metrics}

With three different datasets and tracking tasks, we leverage the most suitable evaluation metric to gauge the performance for each task focusing on different aspects. 

\textbf{CroHD~\cite{sundararaman2021tracking}.} 
For CroHD, we used five different evaluation metrics to comprehensively evaluate the model performance. ATE-Vis and ATE-Occ evaluate the average track error in pixels each time with an $8$-frames video, for targets that stay fully-visible and ones that undergo occlusion, respectively. %The difference between the two is that we include occluders in ATE-Occ for a bigger challenge. 
The other three metrics MTE, $\delta$, and Survival Rate, focus on longer-frame tracking. MTE denotes the median track error in pixels; $\delta$ is an accuracy metric measuring the ratio of points in our predictions that are within a threshold distance of ground truth, \emph{\textbf{averaged}} over thresholds $(1,2,4,8,16)$. Survival measures the average time that a track manages to stay ``on target'' (within a threshold distance 16) without falling off the trajectory and switching to other object or points, measured as a ratio of video length~\cite{zheng2023point}. % which is a typical error in long-time tracking. 

\textbf{Horse30~\cite{mathis2021pretraining}} Similar to CroHD, we include Average Track Error (ATE) and Median Track Error (MTE) for evaluation purpose.

\textbf{Tap-Vid-DAVIS~\cite{doersch2022tap}.}
For Tap-Vid-DAVIS, we use the two evaluation metrics from the benchmark~\cite{doersch2022tap}: $delta$, measuring tracking position accuracy (as done in CroHD), and Average Jaccard (AJ), which is a metric that incorporates both position accuracy and occlusion accuracy.

\textbf{MPI-Sintel~\cite{sintel}.}
For MPI-Sintel, we use the widely-adopted end-point error metric (EPE), which measures the average $\ell_2$ error of the flow estimates.
% crohd
% davis
% sintel
% introduce davis, sintel, crohd, and the different eval metrics

% mention that we measure performance on every video independently

\subsection{Baselines}
In our experiments, we consider the following baselines.

\textbf{DINO~\cite{dino}.} This is a vision transformer (ViT-S \cite{dosovitskiy2020image} with patch size 8) trained with contrastive self-supervision setup on ImageNet~\cite{imagenet_cvpr09}. We use the authors' publicly released code for converting this model into a tracker, which works per-pixel feature matching, with the help of a memory pool~\cite{mast}. 

\textbf{Tap-Net~\cite{doersch2022tap}.} This is a recent method designed specifically for multi-frame point tracking. It estimates pixel correspondences by taking the argmax of frame-by-frame cost maps computed efficiently using time-shifted convolutions~\cite{lin2019tsm}.

\textbf{Self-supervision with color constancy.} This has been previously explored in methods that train from scratch~\cite{yu2016back,jonschkowski2020matters,selflow,stone2021smurf}, but to our knowledge has not been applied to refine pre-trained supervised models. We consider this as a baseline in our work, by fine-tuning RAFT and PIPs with classic color constancy and edge-aware smoothness objectives. 

\begin{table}
  \caption{\textbf{Evaluation in CroHD.} Ours clearly surpasses baselines in most of the metrics and improves ATE of pre-trained PIPs\cite{harley2022particle} by $10\%$. Existing self-supervision methods make performance worse.}
  \label{tab:crohd}
  \centering
  \resizebox{0.5\textwidth}{!}
  {
  \begin{tabular}{lccccc}
    \toprule
    %\multicolumn{2}{c}{Part}                   \\
    %\cmidrule(r){1-2}
    Method & ATE-Vis.~$\downarrow$ & ATE-Occ.~$\downarrow$ & MTE~$\downarrow$ & $\delta$~(\%)~$\uparrow$ & Survival~(\%)~$\uparrow$ \\
    \midrule
    \multicolumn{6}{c}{Supervised Out of Domain }  \\\midrule
    RAFT~\cite{teed2020raft}         & 8.04  & 13.21 & 82.76  & 15.82   & 62.22  \\
    DINO~\cite{dino}                 & 23.04 & 25.91 & 116.80 & 8.46    & 37.11  \\
    PIPs~\cite{harley2022particle}   & 4.57  & 7.71  & 8.35   & \textbf{50.89}   & 82.37   \\
    \midrule
    \multicolumn{6}{c}{Supervised Out of Domain + Self-Supervised}  \\\midrule
    PIPs+Color   & 4.57  & 7.75 & 8.36  & 50.75   & 82.55    \\
    PIPs+Feat.    & 4.52  & 7.70 & 8.37  & 49.76   & 82.90   \\ 
    \rowcolor{lightgray}
    PIPs+Ours     & \textbf{4.02}  & \textbf{6.79}  & \textbf{7.93}   & 49.88   & \textbf{83.96}   \\
    \bottomrule
  \end{tabular}
  }
%\vspace{-10pt}
\end{table}

\subsection{Results}
We present results of our method on each dataset. Overall, our results demonstrate that our method yields
reliable gains over fully-supervised methods, and significantly surpasses the baselines.

\textbf{CroHD~\cite{sundararaman2021tracking}.}
We use PIPs~\cite{harley2022particle} pre-trained on FlyingThings~\cite{flyingthings16} dataset and run CroHD inference during sampling with a spatial resolution of $(768, 1280)$. The 8-frame cycle consistency filter (Eqn.~\ref{eqn:consty_metric}) and multiple-video fine-tuning (Sec.~\ref{subsec:motionref}) are adopted. We choose $2.5$ for the threshold $\tau$ and $3000$ for finetuning iterations $\kappa$. 

% \begin{minipage}{\textwidth}
% \begin{figure*}
% \begin{minipage}[b]{0.48\textwidth}
% \centering
%     \label{tab:davis}
%     \centering
% % \begin{adjustbox}{max width=.6\textwidth}
% \begin{tabular}{lcc}
% \toprule
% % \textbf{{CroHD}} & & {Short-Tracks} & {} & {Long-Tracks}& {}& {}\\ \midrule
% Method & $\delta$($\%$) $\uparrow$ & AJ$\uparrow$ \\ \midrule
% \multicolumn{3}{c}{Supervised Out of Domain (1)}  \\\midrule
%  COTR~\cite{jiang2021cotr} & $51.3$ & $35.4$ \\
%  RAFT~\cite{teed2020raft} & $46.3$ & $79.6$ \\
%  PIPs~\cite{harley2022particle} & $59.4$ & $42.0$ \\
%  TAP-Net~\cite{doersch2022tap} & $53.1$ & $38.4$ \\\midrule
% \multicolumn{3}{c}{(1) + Self-Supervised}  \\\midrule
% PIPs+Ours & $\mathbf{60.0}$ & $\mathbf{42.5}$ \\
% \bottomrule
% \end{tabular}
% \captionof{table}{\textbf{Evaluation in Tap-Vid-DAVIS.} Ours clearly surpasses all baselines in both $\delta$ and AJ with the proposed self-supervised finetuning.}
% \label{tab:davis}
% \end{minipage}
% \hfill
% \begin{minipage}[b]{0.48\textwidth}
% \centering
% % I changed the visualization of this plot
% \includegraphics[width=0.9\linewidth]{figures/davis.pdf}
% \captionof{figure}{Performance change in $\delta$ for each of the video compared with the baseline pre-trained PIPs model. \emph{\textbf{Positive value denotes improvement.}}}
% \label{fig:davis_fig}
% \end{minipage}
% % \hfill
% % \end{minipage}
% \end{figure*}

\begin{table}
\centering
\captionof{table}{\textbf{Evaluation in Tap-Vid-DAVIS.} Ours clearly surpasses all baselines in both $\delta$ and AJ with the proposed self-supervised finetuning.}
\begin{tabular}{lcc}
\toprule
% \textbf{{CroHD}} & & {Short-Tracks} & {} & {Long-Tracks}& {}& {}\\ \midrule
Method & $\delta$($\%$) $\uparrow$ & AJ$\uparrow$ \\ \midrule
\multicolumn{3}{c}{Supervised Out of Domain}  \\\midrule
 COTR~\cite{jiang2021cotr} & $51.3$ & $35.4$ \\
 RAFT~\cite{teed2020raft} & $46.3$ & $79.6$ \\
 PIPs~\cite{harley2022particle} & $59.4$ & $42.0$ \\
 TAP-Net~\cite{doersch2022tap} & $53.1$ & $38.4$ \\\midrule
\multicolumn{3}{c}{Supervised Out of Domain + Self-Supervised}  \\\midrule
\rowcolor{lightgray}
PIPs+Ours & $\mathbf{60.0}$ & $\mathbf{42.5}$ \\
\bottomrule
\end{tabular}
%\vspace{-10pt}
\label{tab:davis}
\end{table}

\begin{figure}[t]
\centering
% I changed the visualization of this plot
\includegraphics[width=1.0\linewidth]{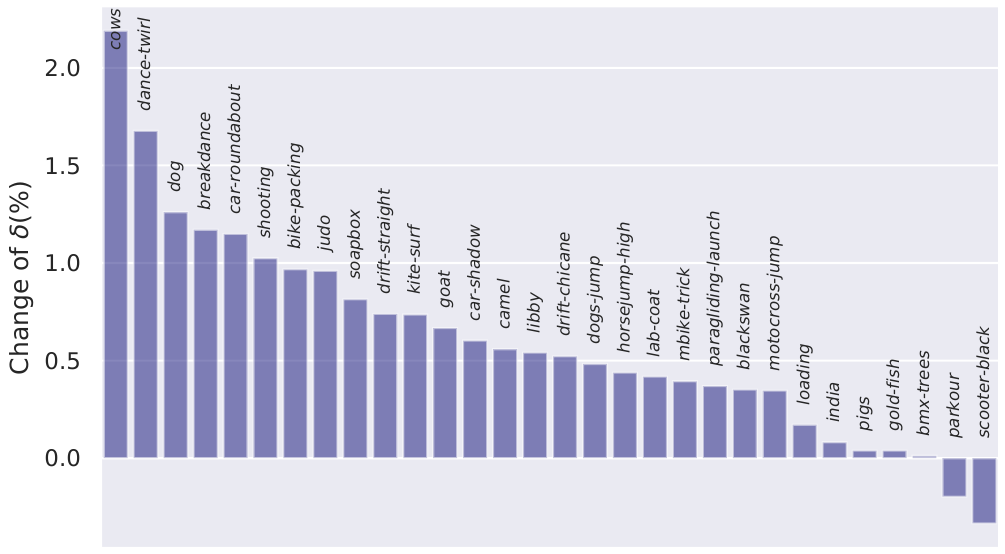}
\captionof{figure}{\textbf{Evaluation in Tap-Vid-DAVIS.} Performance change in $\delta$ for each of the video compared with the baseline pre-trained PIPs model. \emph{{Positive value denotes improvement (i.e., accuracy increasing)}}}
\label{fig:davis_fig}
%\vspace{-6pt}
% \hfill
% \end{minipage}
\end{figure}

% \begin{minipage}{\textwidth}
% \begin{figure*}
%     \begin{minipage}[b]{0.46\textwidth}
%     \centering
%     % \includegraphics[width=\linewidth]{figures/raft_all.pdf}
%     \includegraphics[width=\linewidth]{figures/output.png}
%     \captionof{figure}{Percent change in EPE for each video compared with the pre-trained RAFT~\cite{teed2020raft} model. Negative value denotes improvement (i.e., error being lowered).}
%     \label{fig:raft_all}
%     \end{minipage}
%     \hfill
%     \begin{minipage}[b]{0.46\textwidth}
%     \centering
%     \includegraphics[width=\linewidth]{figures/drawsave.pdf}
%     \captionof{figure}{Comparison of visualizations of optical flows on Sintel produced by the pre-trained RAFT~\cite{teed2020raft} model and RAFT after our self-supervised refinement. Difference regions are highlighted in red boxes. Ours help to refine the flow outputs by cleaning noisy tracks and completing missing objects.}
%     \label{fig:raft_viz}
%     \end{minipage}
% % \end{minipage}
% \end{figure*}
\begin{figure}[t]
    \centering
    \includegraphics[width=\linewidth]{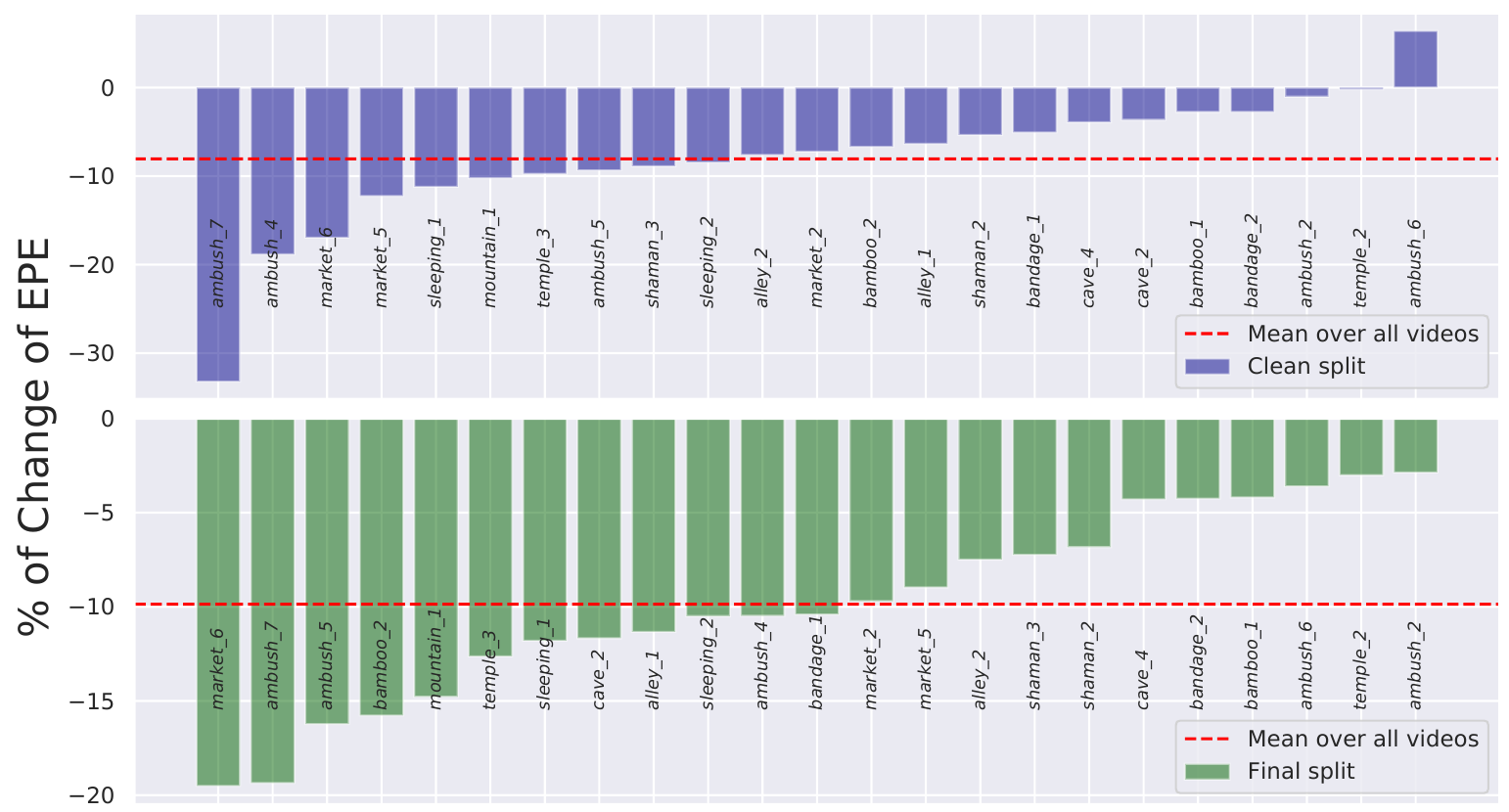}
    \captionof{figure}{\textbf{Evaluation in MPI-Sintel.} Percent change in EPE for each video compared with the pre-trained RAFT~\cite{teed2020raft} model. \emph{{Negative value denotes improvement (i.e., error decreasing).}}}
    \label{fig:raft_all}
   %\vspace{-15pt}
\end{figure}

\begin{figure}[t]
    \centering
    \includegraphics[width=\linewidth]{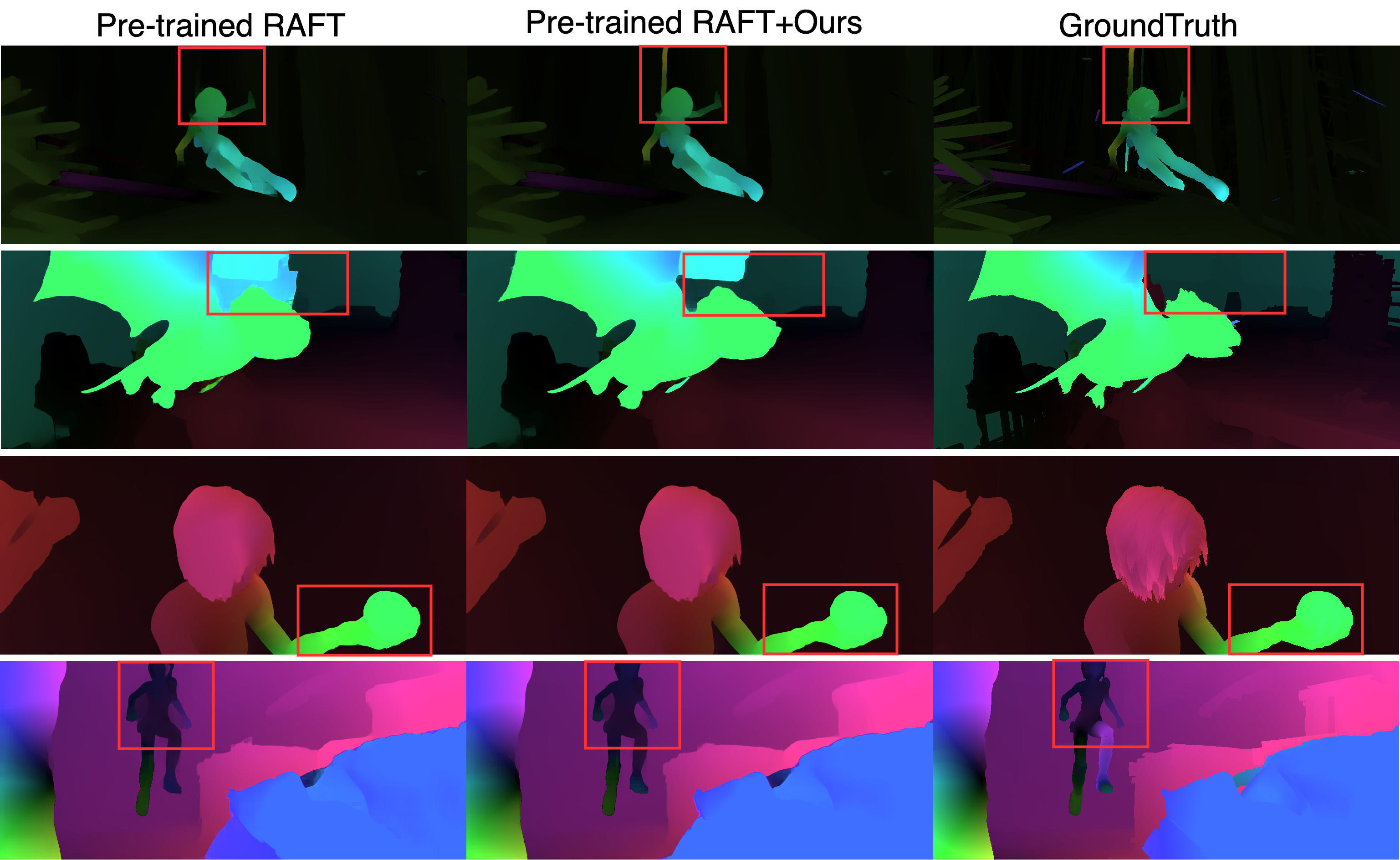}
    \captionof{figure}{Comparison of optical flow visualizations on Sintel produced by the pre-trained RAFT~\cite{teed2020raft} and RAFT with ours, with difference highlighted in red boxes. Ours refine the flows by cleaning noisy tracks and completing missing objects.}
    \label{fig:raft_viz}
% \end{minipage}
%\vspace{-15pt}
\end{figure}

Results and comparisons are provided in Table~\ref{tab:crohd}. We compare with state-of-the-art methods supervised out-of-domain~\cite{teed2020raft, dino, harley2022particle}, and also methods that perform self-supervised finetuning on a pre-trained model like ours. Our method clearly surpasses all of the baselines with better scores on most of the metrics. Notably, compared with pre-trained PIPs where our optimization begins, we reduce ATE-Vis and ATE-Occ from $4.57$ and $7.71$ to $4.02$ and $6.79$, improving by around $10\%$ with our self-supervised refinement procedure. 

For baselines, we also try several self-supervised techniques to finetune the pre-trained model as ours, namely the PIPs+Color and PIPS+Feat shown in Table~\ref{tab:crohd}. The idea of them is to use a self-supervised loss measuring the difference between the color or the image features at each predicted track position. We could observe that both these two techniques are not effective, demonstrating even worse scores on many evaluation metrics than the pre-trained model. This further implies the effectiveness of our proposed procedure in finetuning a pre-trained motion model.

%\begin{table}[ht!]
%    \caption{Performance and comparison of our proposed method with SOTA and several strong baselines. Ours clearly surpasses all baselines and improve ATE of pre-trained PIPs by $10\%$ only with self-supervised training for refinement on the target domain. SoD means Supervised Out of Domain; SS means Self-Supervised.}
%    \label{tab:crohd}
%\begin{adjustbox}{max width=\textwidth}
%\begin{tabular}{@{}l|l|ll|lll@{}}
%\toprule
%\textbf{CroHD} & & Short-Tracks &  & Long-Tracks& & \\ \midrule
%& Method &ATE-VIS$\downarrow$ & ATE-OCC$\downarrow$ & MTE$\downarrow$ & $\delta$($\%$) $\uparrow$& Survival($\%$) $\uparrow$\\ \midrule
% & {Method}      & {EPE$\downarrow$}     & {}   &&&     \\\midrule
%SoD & RAFT    & $8.04$         & $13.21$ &$82.76$ &$15.82$ & $62.22$       \\
% & DINO    & $23.04$         & $25.91$ &$116.80$ &$8.46$ & $37.11$       \\
% & PIPs    & $4.57$      &$7.71$  & $8.35$ &$\mathbf{50.89}$ &$82.37$        \\ \midrule
%SoD+SS & PIPs-pto    & $8.04$         & $13.21$ &$82.76$ &$15.82$ & $62.22$       \\
% & PIPs-feat    & $8.04$         & $13.21$ &$82.76$ &$15.82$ & $62.22$       \\
%\rowcolor{lightgray}
% & \textbf{PIPs+Ours}    & $\mathbf{4.14}$         & $\mathbf{6.99}$ &$\mathbf{7.93}$ &$49.88$ & $\mathbf{83.96}$       \\  \bottomrule
%\end{tabular}
%\end{adjustbox}
%\end{table}

\textbf{Horse30~\cite{mathis2021pretraining}}
We use PIPs~\cite{harley2022particle} pre-trained on FlyingThings~\cite{flyingthings16} dataset and run Horse30 inference during sampling with a spatial resolution of $(256, 448)$. The 8-frame cycle consistency filter (Eqn.~\ref{eqn:consty_metric}) and multiple-video fine-tuning (Sec.~\ref{subsec:motionref}) are adopted. We choose $3$ for the threshold $\tau$ and $6000$ for finetuning iterations $\kappa$. 

Our results on this dataset are shown in Table~\ref{tab:horse}. Compared to the pre-trained PIPs~\cite{harley2022particle}, our method demonstrates a marked improvement in ATE and MTE. Specifically, we observed a decrease in ATE from $12.31$ to $10.81$ and in MTE from $4.40$ to $3.97$ with our self-supervised motion refinement technique. For reference, we also include the performance of pre-trained RAFT~\cite{teed2020raft} model, which registers an ATE of $12.46$ and MTE of $4.03$. We also include the PIPs+Color baseline implemented in the same way as in CroHD to showcase that the traditional motion self-supervision technique fails on pre-trained model. Concretely, it achieves an ATE of $12.15$ and MTE of $4.40$, indicating that it barely helps to improve model performance. These results showcase the effectiveness of our method.

\begin{table}
  \caption{\textbf{Evaluation in Horse30.} Ours clearly surpasses baselines in all of the metrics and improves ATE of pre-trained PIPs\cite{harley2022particle} by more than $10\%$. Existing self-supervision methods make performance worse.}
  \label{tab:horse}
  \centering
  \resizebox{0.35\textwidth}{!}
  {
  \begin{tabular}{lcc}
    \toprule
    %\multicolumn{2}{c}{Part}                   \\
    %\cmidrule(r){1-2}
    Method & ATE~$\downarrow$& MTE~$\downarrow$ \\
    \midrule
    \multicolumn{3}{c}{Supervised Out of Domain}\\\midrule
    RAFT~\cite{teed2020raft}         & $12.46$& $4.03$\\
    PIPs~\cite{harley2022particle}   & $12.31$& $4.63$\\
    \midrule
    \multicolumn{3}{c}{Supervised Out of Domain + Self-Supervised}\\\midrule
    PIPs+Color   & $12.15$& $4.40$\\ 
    \rowcolor{lightgray}
    PIPs+Ours     & $\mathbf{10.81}$& $\mathbf{3.97}$\\
    \bottomrule
  \end{tabular}
  }
%\vspace{-10pt}
\end{table}

\textbf{TAP-Vid-DAVIS~\cite{doersch2022tap}.}
We use PIPs~\cite{harley2022particle} pre-trained on FlyingThings~\cite{flyingthings16} dataset and run DAVIS inference during sampling with a spatial resolution of (288, 512). The 8-
frame cycle consistency filter (Eqn.~\ref{eqn:consty_metric}) and single-video finetuning (Sec.~\ref{subsec:motionref}) are adopted. Specifically, we choose
$1$ for the threshold $\tau$ and 100 for finetuning iterations $\kappa$ for each video. For a single-video finetuning setup, we show performance change on each individual video with the baseline pre-trained model. To compare with other methods on the same scale, we also report an average of our performance over all videos.
%\begin{table}[t]

Results and comparisons are provided in Table~\ref{tab:davis} and Fig.~\ref{fig:davis_fig}. We compare with state-of-the-art methods~\cite{jiang2021cotr, teed2020raft, harley2022particle, doersch2022tap} supervised out of domain including the latest works like Tap-Net~\cite{doersch2022tap}. We beat all of the baselines with better $\delta$ and AJ. Notably, compared with PIPs where we start from, we improve $\delta$ and AJ from $59.4$ and $42.0$ to $60.0$ and $42.5$. Moreover, we also show the performance change in $\delta_\text{avg}$ for each video in Fig.~\ref{fig:davis_fig}. We could observe positive change through our self-supervised refinement on almost all of the 30 videos except two videos `parkour' and `scooter-black'. %In some of the videos like `cows', improvement is even more obvious. 

Compared with the CroHD results in Table~\ref{tab:crohd} where we improve the pre-trained model by $10\%$, results on Tap-Vid-DAVIS do not show as much improvement. However, we note that each Tap-Vid-DAVIS video is only $5$ seconds long on average (not much space to extract pseudo-labels), and ground truth is only available for a small number of points (less than 30 per video). %nd extremely sparse ground truth, we think the improvements seen in Table~\ref{tab:davis} and Fig.~\ref{fig:davis_fig} are already reasonably decent.
%\begin{table}[ht!]

% \begin{wrapfigure}{r}{0.45\textwidth}
\begin{table}
\captionof{table}{\textbf{Evaluation in MPI-Sintel.} A clear advantage is observed for our method, improving pre-trained RAFT by around $10\%$. Existing self-supervised methods do not help.}
\label{tab:raft}
\centering
\begin{tabular}{lll}
\toprule
Method      & EPE$\downarrow$     & {}        \\\midrule
 & Clean & Final \\ \midrule
    \multicolumn{3}{c}{Supervised Out of Domain}  \\\midrule
 FlowNet2~\cite{flownet2}    & $2.02$         & $3.14$         \\
      PWC-Net~\cite{sun2018pwc}     & $2.55$         & $3.93$         \\
      VCN~\cite{yang2019volumetric}         & $2.21$         & $3.62$         \\
  RAFT~\cite{teed2020raft}        & $1.46$         & $2.72$         \\ \midrule
    \multicolumn{3}{c}{Self-Supervised}  \\\midrule
 SelFlow~\cite{selflow}     & $2.88$         & $3.87$         \\
  UFlow~\cite{jonschkowski2020matters}       & $3.01$         & $4.09$         \\
  SMURF-test~\cite{stone2021smurf}  & $1.99$         & $2.80$         \\ \midrule
    \multicolumn{3}{c}{Supervised Out of Domain + Self-Supervised}  \\\midrule
 {RAFT+Color} & $1.48$         & $2.73$ \\  
       {RAFT+Col.+Smooth} & $1.47$         & $2.74$ \\ 
       {RAFT+Col.+Edge-Sm.} & $1.51$         & $2.84$ \\
       \rowcolor{lightgray}
      {RAFT+Ours} &     $\mathbf{1.32}$         &    $\mathbf{2.46}$ \\        
      \bottomrule
\end{tabular}
%\vspace{-10pt}
\end{table}
% \end{wrapfigure}

\textbf{MPI-Sintel.}
For MPI-Sintel, we use RAFT~\cite{teed2020raft} pre-trained on FlyingThings datasets and run Sintel inference during sampling with a spatial resolution of (436, 1024). We use the 2-frame cycle consistency filter (Eqn.~\ref{eqn:raftconsty}) augmented by color consistency (Eqn.~\ref{eqn:raftcolorconsty}) and fine-tune on each video individually. % single-video finetuning (Sec.~\ref{subsec:motionref}) are adopted. 
We use the threshold hyperparameters $\alpha=0.005$, $\beta=0.25$, and $\gamma=0.1$. We choose $100$ for finetuning iterations
$\kappa$ for each video. %We provide ablations later on how much the augmented color-consistency-aware filter can boost performance compared with pure cycle-consistency-based filter for optical flow.
%SImilarly as TAP-Vid-DAVIS, 
We evaluate performance change on each individual scene, and also measure the average across videos. % the baseline pre-trained RAFT but also include an average over all videos to compare with others.

Quantitative comparisons are provided in Table~\ref{tab:raft} and Fig.~\ref{fig:raft_all}, and qualitative comparisons are in Fig.~\ref{fig:raft_viz}. We compare with state-of-the-art methods supervised out-of-domain, self-supervised methods, and self-supervised techniques to finetune the pre-trained model. We surpass all of the baselines, with better 
EPE on both clean and final splits of MPI-Sintel. Notably, compared with the pre-trained RAFT model, we reduce the EPE for both clean and final from $1.46$ and $2.72$ to $1.32$ and $2.46$, demonstrating around $10\%$ improvement. Since evaluation variance is quite huge on Sintel, we visualize the percentage EPE change for each of the videos in Figure~\ref{fig:raft_all}. We can clearly observe that our self-supervised refinement improves the model performance on all of the scenes except `ambush\_6' in the clean split. On the other scenes, we observe obvious and significant improvements, even $30\%$ on scenes like `ambush\_7'. We visualize the optical flow results in Fig.~\ref{fig:raft_viz}, and highlight the noticeably different regions in red boxes. We observe that our proposed method helps refine the optical flow output by completing missing objects, and improving smoothness in large regions.

As baselines to our optimization strategy, we try using color constancy as a self-supervised loss to finetune the pre-trained model, optionally combined with a smoothness regularizer or edge-aware smoothness regularizer. We list these in Table~\ref{tab:raft} as RAFT+Color, RAFT+Col.+Smooth and RAFT+Col.+Edge-Sm. In practice, we found that none of these techniques help, and EPE simply rises as training progresses. %As seen in Table~\ref{tab:raft}, those methods worsen the performance of pre-trained RAFT. Our own self-supervision reliably improves results. 

% \begin{table}[ht!]

% \begin{figure}[ht!]
% \begin{minipage}{\linewidth}
% \begin{minipage}[b]{0.4\linewidth}
% % \begin{table}
% \begin{adjustbox}{max width=\textwidth}
% \begin{tabular}{l|lll}
% \caption{Performance and comparison of our proposed method with SOTA and several strong baselines.}
% \toprule
% \textbf{Sintel-Train} & & Clean & Final \\ \midrule
% & Method      & EPE$\downarrow$     & {}        \\\midrule
% SoD & FlowNet2    & $2.02$         & $3.14$         \\
%      & PWC-Net     & $2.55$         & $3.93$         \\
%      & {VCN}         & $2.21$         & $3.62$         \\
%  & {RAFT}        & $1.46$         & $2.72$         \\ \midrule
% SS & SelFlow     & $2.88$         & $3.87$         \\
%  & UFlow       & $3.01$         & $4.09$         \\
%  & SMURF-test  & $1.99$         & $2.80$         \\ \midrule

%  SoD + SS     & {RAFT+Cr} & $1.48$         & $2.73$ \\  
%       & {RAFT+Cr+Sth} & $1.47$         & $2.74$ \\ 
%       & {RAFT+Cr+EdgSth} & $1.51$         & $2.84$ \\ 
%     \rowcolor{lightgray}
%      & {RAFT + Ours} &     $\mathbf{1.32}$         &    $\mathbf{2.46}$ \\        
% \end{tabular}
% \end{adjustbox}
% \end{minipage}%
% \hfill
% \begin{minipage}[b]{0.6\linewidth}
%     % % \centering
%     % \includegraphics[width=\linewidth]{figures/raft_all.pdf}
%     % \caption{Performance change in EPE for each of the video compared with the baseline pre-trained RAFT model. Negative value denotes improvement.}
%     % \label{fig:raft_all}
% \end{minipage}
% % \end{table}
% \end{minipage}
% \end{figure}

\subsection{Ablation Studies}
\begin{figure}[t]
    \centering
    \includegraphics[width=\linewidth]{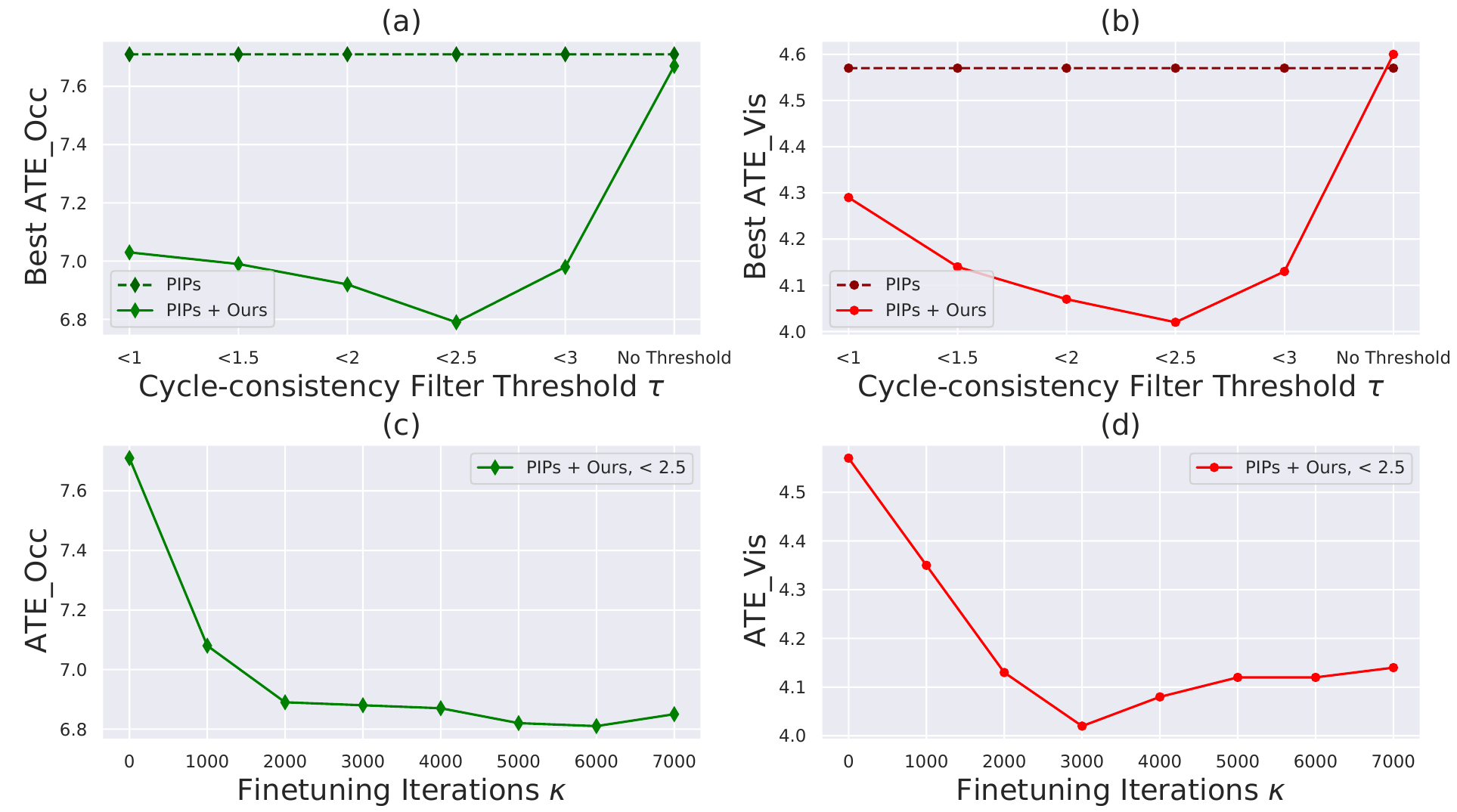}
    \caption{\textbf{Ablation study in CroHD.} We show ATE\_VIS and ATE\_OCC for picking pseudo-labeled tracks with different threshold $\tau$ (a and b) and fine-tuned for different iterations $\kappa$ (c and d). All metrics are the smaller the better. Optimal performance is observed at $\tau$ of $2.5$ and $\kappa$ of $3000$.}
    \label{fig:crohd_ablation}
    %\vspace{-15pt}
    \vspace{1em}
\end{figure}

We conduct two ablation studies to investigate the effect of the design choices and hyperparameters. First, we study the effect of the threshold $\tau$ and finetuning iterations $\kappa$. Results are shown in Fig.~\ref{fig:crohd_ablation}. We conduct this study on CroHD. We collect the best ATE\_OCC and ATE\_VIS for various $\tau$s, and as an additional variant, randomly sample the tracks to choose the pseudo-label finetuning set, denoted as `No Threshold' in the figure. % Fig.~\ref{fig:crohd_ablation}. As seen in the figure, 
We find that random sampling leads to almost no performance change, while the cycle-consistency-based filtering leads to improvements, across various values of $\tau$. We observe that the optimal performance is achieved at $\tau$ of $2.5$ in this dataset, with a U-shaped curve around this value. %degradation in performance with either smaller or larger $\tau$. 
This follows expectations, since a loose filter (with large $\tau$) includes too many noisy motion estimates, which corrupts the pre-trained weights; a strict filter (with small $\tau$) contains too few good estimates to provide useful guidance to the pre-trained model for the new context.

Fig.~\ref{fig:crohd_ablation}-(c) and Fig.~\ref{fig:crohd_ablation}-(d) measure the effect of fine-tuning iterations $\kappa$. The plot illustrates that $\kappa=3000$ achieves the best performance. Performance degrades slightly with either smaller or larger $\kappa$, which either insufficiently finetunes the model on the pseudo-label set, or leads to slight overfitting.

Table~\ref{tab:raft-ablation} measures the effectiveness of different strategies for choosing pseudo-labels for optical flow.  %of the color constancy metric we use for sampling good optical flow pseudo-tracks. 
We conduct this study in Sintel~\cite{sintel}. 
%Results are provided in . As seen, 
Both color constancy and cycle consistency alone are helpful, with cycle consistency slightly outperforming color constancy. Combining the two filtering strategies yields the best performance. 

\begin{table}[t]
  \centering
  \caption{\textbf{Ablation study in MPI-Sintel-Clean.} Results are evaluated on $8$ videos from the Sintel-Clean train split. Both Cycle Consistency and Color Constancy self-supervised metrics lead to improvements from the pre-trained RAFT. Combining both metrics works the best.
  }
  \label{tab:raft-ablation}
  
  \begin{tabular}{llll}
    \toprule
    Method     &Color Constancy &Cycle Consistency & EPE$\downarrow$             \\\midrule
    RAFT     &n/a & n/a   & $1.05$                 \\ \midrule
    RAFT + Ours   & \cmark  & \xmark  & $0.98$        \\
    RAFT + Ours   & \xmark  & \cmark  & $0.99$       \\
 %    \rowcolor{lightgray}
    RAFT + Ours   & \cmark  & \cmark  & $\mathbf{0.93}$       \\
    
    % RAFT+Ours        & $1.46$         & $2.72$         \\
    % RAFT+Ours        & $1.46$         & $2.72$         \\ \bottomrule
    \bottomrule
    \end{tabular}
%\vspace{-10pt}
\end{table}

% \subsection{Limitations}
% Though our method brings solid improvements on different datasets for various tracking tasks, there are some limitations as well. On few videos like `scooter-black' of Tap-Vid-DAVIS and `ambush\_6' of MPI-Sintel clean, performance gets slightly worse by finetuning with our pseudotracks. Right now, we do not have a reliable metric to tell when this scenario would happen and prevent it beforehand. 
\section{Discussion and conclusion}

In this paper, we present a %novel, simple, and effective 
procedure to improve state-of-the-art supervised motion models with only self-supervised fine-tuning. 
Our %entire 
procedure is a %simple 
two-stage method where we first use a pre-trained motion model to sample and filter cycle-consistent tracks in the test domain, %followed by finetuning on augmented versions of them. 
and then we fine-tune on these tracks, with added augmentations. 
%We conduct a comprehensive empirical study, 
We compare our method with state-of-the-art supervised models, self-supervised training methods, and self-supervised finetuning techniques. 
We find our method improves the pre-trained model performance, on three different datasets, with two different tracking models, % with different tracking tasks,
whereas %the 
existing %motion 
self-supervised techniques %fail or 
worsen %the 
performance. We hope our method will inspire %more 
future work %s into
on refining pre-trained motion models. 
\section{Acknowledgments.}
 This work was supported by the Toyota Research Institute under the University 2.0 program, ARL grant W911NF-21-2-0104, and a Vannevar Bush Faculty Fellowship. The authors thank Leonid Keselman for suggesting the title.

{\small
\bibliographystyle{IEEEtran}
\bibliography{99_refs}
}

\section{Supplementary Material}

\subsection{Qualitative Comparisons}
We provide additional visualization results of our refined optical flows and tracks to illustrate the improvements brought by our method. Performance improvements of PIPs are better demonstrated with videos, and we have GIFs comparing ours with PIPs on our project website. We encourage readers to take a look at the videos to gain a better understanding of our qualitative improvement.
\begin{figure*}
    \centering
    \includegraphics[width=\linewidth]{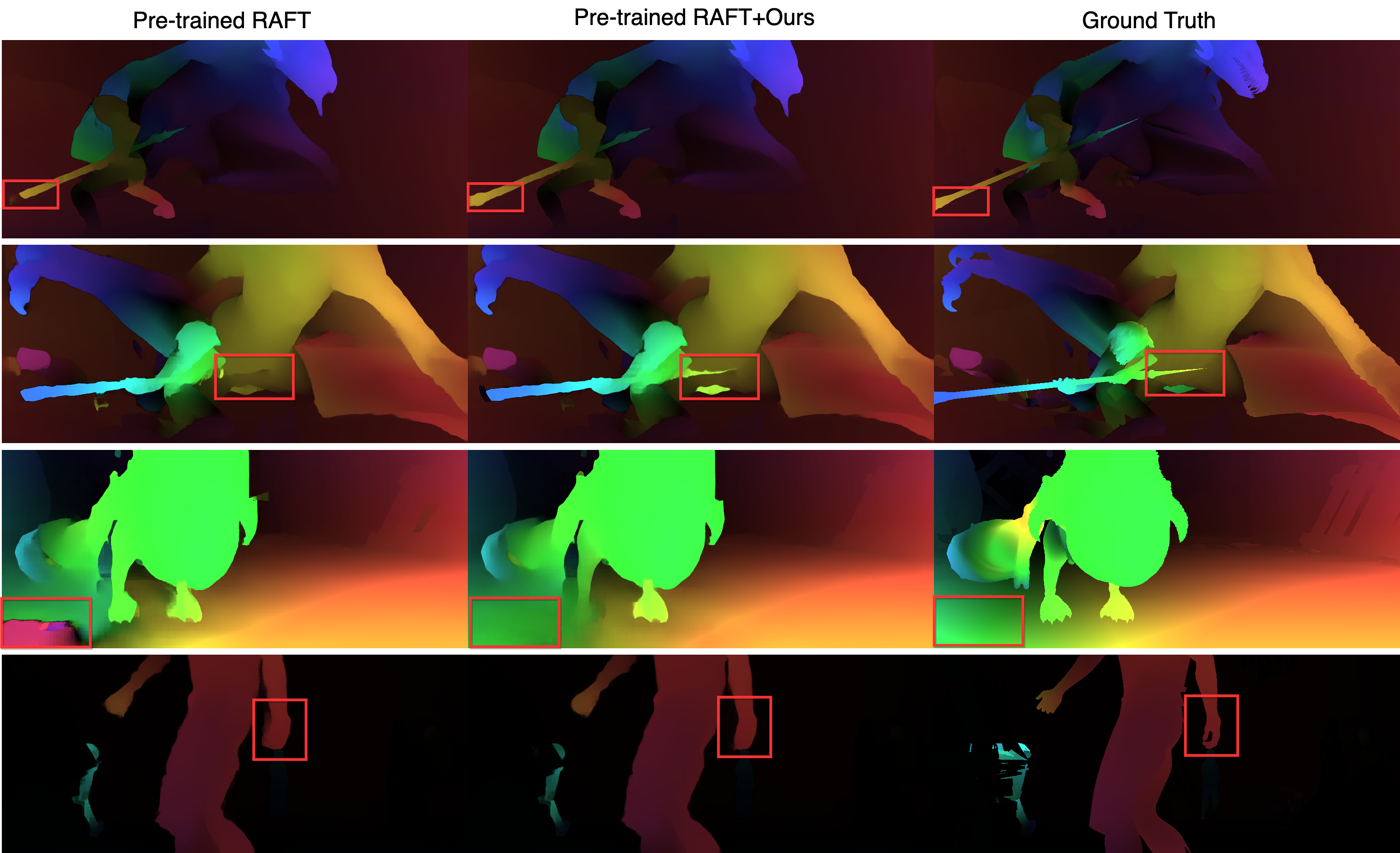}
    \includegraphics[width=\linewidth]{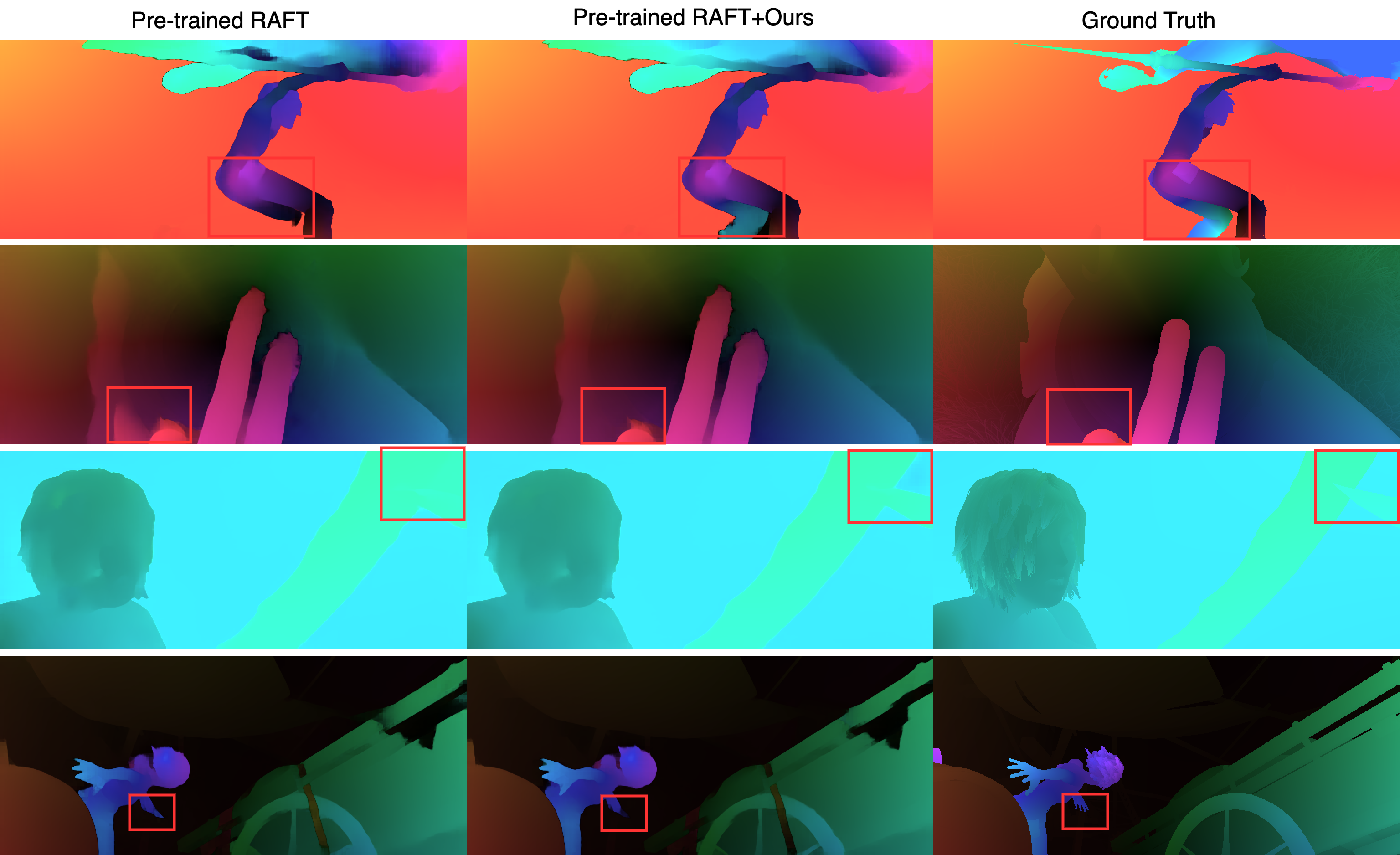}
    \caption{Comparison of visualizations of optical flow on MPI-Sintel produced by the pre-trained RAFT~\cite{teed2020raft} model and RAFT after our self-supervised refinement. Difference regions are highlighted in red boxes. Ours help to refine the flow outputs by cleaning noisy flows, completing missing objects, and improving smoothness.}
    \label{fig:more_flow_viz}
\end{figure*}

\begin{figure*}
    \centering
    \includegraphics[width=.85\linewidth]{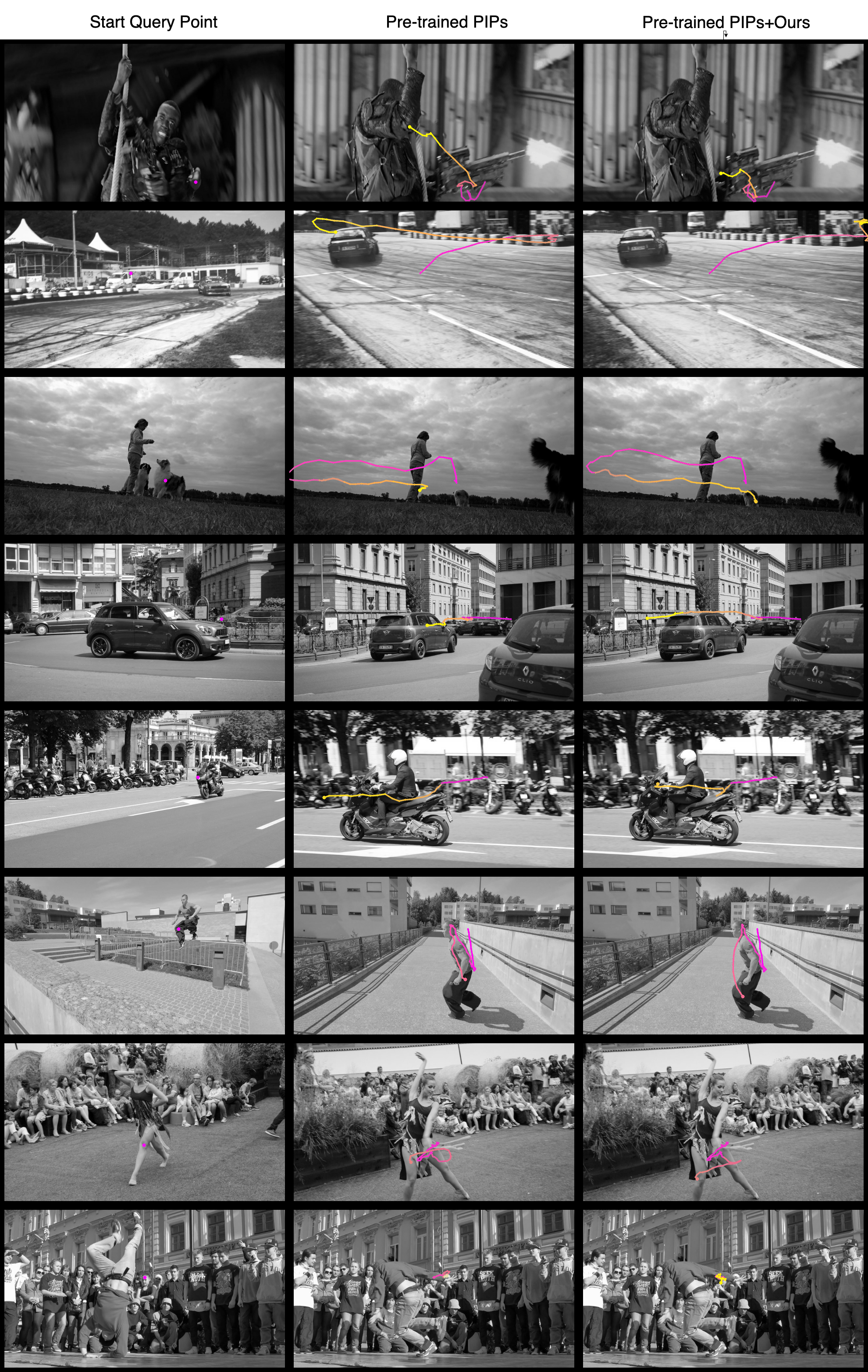}
    \caption{Comparison of visualizations of tracks on Tap-Vid-DAVIS produced by the pre-trained PIPs~\cite{harley2022particle} model and PIPs after our self-supervised refinement. We visualize the query point location on the first frame and tracks to the end of the video. We improve PIPs by producing more accurate tracks.}
    \label{fig:pips_viz}
\end{figure*}

\subsection{Implementation Details}
In this section, we provide additional implementation details, to aid reproducibility on all three tasks. We use PyTorch as our deep learning framework and train on Titan Xp GPUs. We will also release our code, generated pseudo labels, and models. % upon acceptance.

\textbf{CroHD.}
Our base model is PIPs with $S=8$, which means that we train motion tracks for 8 frames simultaneously. Another hyperparameter of PIPs training is $N$ which decides how many tracks we train each time on an 8 frames chunk. On CroHD finetuning, we use $N = 128$. We also pick training spatial resolution as $(512, 852)$ which is the maximum image size we can support on our GPU during training. Batch size is set as $1$ but we set a gradient accumulation period of $8$, which means we perform an optimizer step only after backward passes of $8$ samples to simulate a larger batchsize. In terms of optimization settings, we use AdamW optimizer with an initial learning rate set as $1e-5$, wdecay as $1e-4$, and epsilon as $1e-8$. The learning rate is decayed overtimes by a CosineAnnealing scheduler. We finetune for $3000$ iterations in total on our generated pseudo labels.

\textbf{Horse30.}
Similar to CroHD, our base model is PIPs with $S=8$ and we use $N=128$. We train with a spatial resolution of $(256, 448)$. The batch size is set as $1$ but we use a gradient accumulation period of $8$ to simulate a larger batchsize. In terms of optimization settings, we use AdamW optimizer with an initial learning rate set as $1e-5$, wdecay as $1e-4$, and epsilon as $1e-8$. The learning rate is decayed overtimes by a CosineAnnealing scheduler. We finetune for $6000$ iterations in total on our generated pseudo labels. 

\textbf{Tap-Vid-DAVIS.}
Similar to CroHD, our base model is PIPs with $S=8$ and we use $N=128$. We train with a spatial resolution of $(288, 512)$. The batch size is set as $1$ but we perform horizontal and vertical flipping similar to PIPs during training to augment the batch size to $4$. Again, we set a gradient accumulation period of $16$ to simulate a larger batchsize. In terms of optimization settings, we use AdamW optimizer with an initial learning rate set as $5e-7$, wdecay as $1e-4$, and epsilon as $1e-8$. The learning rate is decayed overtimes by a CosineAnnealing scheduler. The learning rate may appear too small for the model to actually learn anything, but even the initial learning rate used in PIPs is only $1e-4$. We experimented and found that larger learning rate will mess up the good pretrained PIPs weights. We finetune for $1600$ iterations in total on our generated pseudo labels on each video independently. 

\textbf{MPI-Sintel.}
Our base model is RAFT with the number of inner refinement iterations as $32$. We train on the spatial resolution of $(256, 512)$. The batch size is set as $12$. In terms of optimization settings, we use AdamW optimizer with an initial learning rate set as $1e-4$, wdecay as $1e-4$, and epsilon as $1e-8$. The learning rate is decayed overtimes by a CosineAnnealing scheduler. We finetune for $1000$ iterations in total on our generated pseudo labels on each scene independently. 

\subsection{Evaluation Metrics}
In this section, we provide additional details on the computation of our evaluation metrics.

\textbf{CroHD~\cite{sundararaman2021tracking} and Horse30~\cite{mathis2021pretraining}} 
\begin{itemize}
\item \textbf{ATE.}
On each sample, suppose the 8-frames groundtruth tracks is $y \in \mathbb{R}^{8\times 2}$, and our predicted tracks is $\hat{y} \in \mathbb{R}^{8\times 2}$. Furthermore, suppose we have a valid binary vector $\psi \in \{0, 1\}^{8}$ denoting the validity of data at each time frame. Track error is computed as:
% \begin{equation}
%     ATE = \sum_{t=0}^8 (\|y_t - \hat{y}_t\|_2)
% \end{equation}
\begin{align}
    TE = \frac{1}{8}\sum_{t=0}^8 (\|y_t - \hat{y}_t\|_2) \cdot \psi_t
\end{align}
During validation, we take the average of $TE$ over all validation samples.

\item \textbf{MTE.}
Given the above computation of $TE$, we rank all measurements over validation samples and take the median value as MTE.

\item $\mathbf{\delta}$.
More computation detail of this metric is provided in the Tap-Vid paper \cite{doersch2022tap}. First, define $\delta^\tau$ as the position accuracy for visible points, measured as the fraction of points that are within threshold $\tau$ of groundtruth. Formally, over an 8-frames chunk, this can be defined as:
\begin{align}
    \delta^\tau = \frac{|\bigcup_{t}^8\{1; \|y_t - \hat{y}_t\|_2 < \tau\}|}{8}
\end{align}
Then, for our final metric, we take the average across five different thresholds $\tau$, which are 1, 2, 4, 8, and 16.

\item \textbf{Survival Rate.}
In long-range tracking, a typical tracking error is the tracking point falling off, defined by us as moving more than $50$ pixels away from the ground truth point. For example, out of an $8$ frames chunk, if the tracking point falls off at $t$th frame, its survival rate will be $(t-1)/8$. We compute the average survival rate over all validation samples.

\end{itemize}

\textbf{Tap-Vid-DAVIS~\cite{doersch2022tap}}
\begin{itemize}
\item $\mathbf{\delta}$.
Computation is exactly the same as $\mathbf{\delta}$ for CroHD.

\item \textbf{AJ.}
More computation detail of this metric is provided in the Tap-Vid paper \cite{doersch2022tap}. Jaccard evaluates both occlusion and position accuracy. True positives are defined as points within the threshold $\tau$ of any \emph{visible} ground truth points. False positives are defined as points that are predicted visible, but the ground truth is either occluded or farther than the threshold $\tau$. Jaccard at $\tau$ is the fraction of true positive over true positive plus false positive. Similar to $\delta$, we average over five threshold values: 1, 2, 4, 8, 16.
\end{itemize}

\textbf{MPI-Sintel~\cite{sintel}}
\begin{itemize}
\item \textbf{EPE.}
%More computation detail of the metric can be found in RAFT~\cite{teed2020raft}. 
This is simply measuring the $\ell_2$ distance between the predicted flow destination point and ground truth destination point. Suppose the groundtruth flow vector is $f \in \mathbb{R}^{2}$, giving $x, y$ displacements. Suppose the predicted flow vector is $\hat{f} \in \mathbb{R}^{2}$. EPE is thus computed as:
\begin{align}
    EPE = \|f_1 - \hat{f}_1\|_2
\end{align}
We average the EPE measurements over all samples of the validation split.
\end{itemize}

\subsection{Baselines}
In our main paper, we compare ours with several self-supervised techniques applied to pre-trained motion models similar to ours. We showed that none of those existing straightforward techniques achieve success in finetuning a pretrained motion model in a self-supervised manner. Here, we will provide more computation and reproduction detail of each of the techniques.

\subsubsection{CroHD and Horse30}
\textbf{PIPs+Color.}
This is PIPs+Color in Table \ref{tab:crohd}. Suppose the 8-frames groundtruth tracks is $y \in \mathbb{R}^{8\times 2}$, and our predicted tracks is $\hat{y} \in \mathbb{R}^{8\times 2}$. Moreover, suppose our input RGB images are $X \in \mathbb{R}^{8\times 3 \times H \times W}$ which stacks RGB images of all the temporal frames. Following the standard and classical optical flow definition, we would expect the points on one 8-frames track share similar colors. Also, define $X_t[y_t] \in \mathbb{R}^{3}$ as a bilinear sampling operation computing the RGB value of $X_t$ at position $y_t$. This self-supervised loss can be computed as:
\begin{align}
    \label{eqn:colorloss}
    \textrm{loss}_{\textrm{color}} = \sum_{t=2}^{8} \|X_t[y_t] - X_1[y_1]\|_2     
\end{align}

\textbf{PIPs+Feature.}
This is PIPs+Feat in Table \ref{tab:crohd}. It is similar to PIPs+Color, but when computing the loss, we replace the RGB images $X$ with feature maps $\mathcal{F}$ obtained by the model backbone, to make the matching more robust to lighting and texture effects. %capture more structural detail.

\subsubsection{MPI-Sintel}
\textbf{RAFT+Color.}
This is RAFT+Color in Table \ref{tab:raft}. It is similar to PIPs+Color described above with two differences: (1) we only consider $2$ instead of $8$ frames for optical flow (2) optical flow is dense over the entire image instead of sparse query points. Due to these two reasons, we use the target image and predicted flow vector to reconstruct the source image with backward and compute the $\ell_2$ reconstruction error. This can be considered as Eqn.\ref{eqn:colorloss} with $t=2$ and averaging over the entire image.

\textbf{RAFT+Color+Smoothness.}
This is RAFT+Col.+Smooth in Table \ref{tab:raft}. We also try adding an additional smoothness loss to encourage optical flow vector in the same object to be closer to each other. Given the dense predicted flow over the image as $\hat{F} \in \mathbb{R}^{H \times W \times 2}$, we first compute the spatial gradient $\Delta F$ as $\Delta F = \frac{\partial}{\partial x} \frac{\partial}{\partial y} F$, indicating the spatial change of optical flow. Then, we compute the smoothness loss simply as:
\begin{align}
    \textrm{loss}_{\textrm{smooth}} = \|\Delta F\|_2
\end{align}

\textbf{RAFT+Color+Edge-aware Smoothness.}
This is RAFT+Col+Edge-Sm. in Table \ref{tab:raft}. It is similar to RAFT+Color+Smoothness, but with the smoothness loss downweighted at pixels with large image gradients. % The difference is that we downweight the smoothness loss at image gradients instead of standard smoothness, we also consider the edges appearing in RGB images where the gradients are naturally quite large. 

\end{document}